\documentclass[sigconf]{acmart}
\usepackage{booktabs} 
\usepackage{multirow}
\usepackage{tabu}
\usepackage{makecell}
\usepackage{array}
\usepackage{lipsum}
\usepackage{epstopdf}
\usepackage{graphicx}
\usepackage{epstopdf}
\usepackage{makecell}
\usepackage{booktabs}
\usepackage{multirow}
\usepackage{algorithm}
\usepackage{algorithmic}
\usepackage{subfigure}
\usepackage{enumitem}
\usepackage{enumerate}
\usepackage{amsthm}
\usepackage{soul}

\AtBeginDocument{%
  \providecommand\BibTeX{{%
    \normalfont B\kern-0.5em{\scshape i\kern-0.25em b}\kern-0.8em\TeX}}}

\copyrightyear{2022}
\acmYear{2022}
\setcopyright{acmcopyright}\acmConference[WWW '22 Companion]{Companion Proceedings of the Web Conference 2022}{April 25--29, 2022}{Virtual Event, Lyon, France}
\acmBooktitle{Companion Proceedings of the Web Conference 2022 (WWW '22 Companion), April 25--29, 2022, Virtual Event, Lyon, France}
\acmPrice{15.00}
\acmDOI{10.1145/3487553.3524248}
\acmISBN{978-1-4503-9130-6/22/04}

\begin{document}
\title{Personal Attribute Prediction from Conversations}
\newcommand\blfootnote[1]{%
	\begingroup
	\renewcommand\thefootnote{}\footnote{#1}%
	\addtocounter{footnote}{-1}%
	\endgroup
}
\author{Yinan Liu, Hu Chen, Wei Shen$^*$}
\affiliation{
	\country{TKLNDST, College of Computer Science, Nankai University, Tianjin, 300350, China}
}
\email{{liuyn, 2120210473}@mail.nankai.edu.cn,  shenwei@nankai.edu.cn}



\begin{abstract}
  Personal knowledge bases (PKBs) are critical to many applications, such as Web-based chatbots and personalized recommendation. Conversations containing rich personal knowledge can be regarded as a main source to populate the PKB. Given a user, a user attribute, and user utterances from a conversational system, we aim to predict the personal attribute value for the user, which is helpful for the enrichment of PKBs. However, there are three issues existing in previous studies: 
  (1) manually labeled utterances are required for model training; (2) personal attribute knowledge embedded in both utterances and external resources is underutilized; (3)
  the performance on predicting some difficult personal attributes is unsatisfactory. In this paper, we propose a framework DSCGN based on the pre-trained language model with a noise-robust loss function to predict personal attributes from conversations without requiring any labeled utterances. We yield two categories of supervision, i.e., document-level supervision via a distant supervision strategy and contextualized word-level supervision via a label guessing method, by mining the personal attribute knowledge embedded in both unlabeled utterances and external resources to fine-tune the language model. Extensive experiments over two real-world data sets (i.e., a profession data set and a hobby data set) show  
  our framework obtains the best performance compared with all the twelve baselines in terms of nDCG and MRR.
\end{abstract}



\maketitle
\blfootnote{$^*$Corresponding author}

\section{Introduction}

Some recent studies \cite{balog2019personal,yen2019personal} propose to construct the personal knowledge base (PKB) for individuals, which contains a resource of structured information about entities personally related to
its user, their attributes, and the relations between them. The advantages of the PKB are twofold. On one hand, a PKB can be merged with large general KBs (e.g., Freebase, DBpedia, and YAGO), so that the personal life events are connected with world knowledge. On the other hand, the PKB can provide personal background knowledge for many downstream applications
such as Web-based chatbots, personalized recommendation, and personalized search.

Conversational systems are becoming very important as many personal assistants emerge (e.g., Siri and Cortana). Understanding user utterances plays an essential role in holding meaningful conversations with users. In reality, the PKB is beneficial for understanding user utterances better.
Meantime, conversation data can also be regarded as a main source to populate a PKB by drawing personal attribute knowledge (e.g., hobbies, professions, and visited cities) from the users' conversations in many platforms (e.g., social media and dialogues). To enrich the PKB using personal knowledge from conversations, predicting personal attributes from conversations is an important task that needs to be solved urgently.


Given a subject (i.e., a user of a conversational system), a predicate (i.e., a user attribute), and user utterances, our task is to predict the ranking of the object (i.e., attribute value) for the given subject-predicate combination according to user utterances. For instance, we could rank the student and doctor professions high when the user mentions the words \textit{undergrad}, \textit{school}, \textit{study}, \textit{medicine}, etc. Subsequently, the generated subject-predicate-object triple could be used to populate the PKB.


Recently, several supervised models \cite{tigunova2019listening,tigunova2020charm} have been proposed to predict personal attributes from conversations. The first work \cite{tigunova2019listening} leveraged neural networks based on utterance embeddings and attention mechanisms to solve this task. Tigunova et al. \cite{tigunova2020charm} achieved better performance
by designing a reinforce policy to integrate utterance keyword extraction and Web document retrieval.
However, there exist three issues in previous works:
(1) the construction of training data via manually annotating utterances is required and difficult; (2) personal attribute knowledge embedded in both utterances and external resources (e.g., Wikipedia) is underutilized; (3) the performance on predicting some difficult personal attributes (e.g., profession and hobby) with too many candidate values is unsatisfactory.


To deal with these issues, we propose a novel end-to-end language model based framework DSCGN without requiring any labeled utterances, which derives two categories of supervision, i.e., document-level supervision via a \textbf{\underline{D}}istant \textbf{\underline{S}}upervision strategy and \textbf{\underline{C}}ontextualized word-level supervision via a label \textbf{\underline{G}}uessing method by mining personal attribute knowledge embedded in unlabeled utterances and external resources. DSCGN performs confidence based pre-trained language model (PLM) fine-tuning via a \textbf{\underline{N}}oise-robust loss function incorporating these two categories of supervision.
Our contributions can be summarized as follows:
\setlist[itemize]{leftmargin=0mm,topsep=0mm,parsep=0mm,listparindent=0mm,itemindent=\parindent}

\begin{itemize}[leftmargin=*]
	\setlength{\itemsep}{0pt}
	\setlength{\parsep}{0pt}
	\setlength{\parskip}{0pt}
	\setlength\parindent{0pt}
\item We propose the first language model based framework without requiring any labeled utterances to predict personal attributes from conversations. 

\item We propose a noise-robust loss function that can incorporate document-level supervision and contextualized word-level supervision to fine-tune the language model.
\item Extensive experiments over two real-world data sets (i.e., a profession data set and a hobby data set), which are derived from Reddit show our framework obtains the best performance compared with all the twelve baselines in terms of nDCG and MRR.
\end{itemize}

\begin{figure}[!t]
	\centering
	\includegraphics[width=0.48\textwidth]{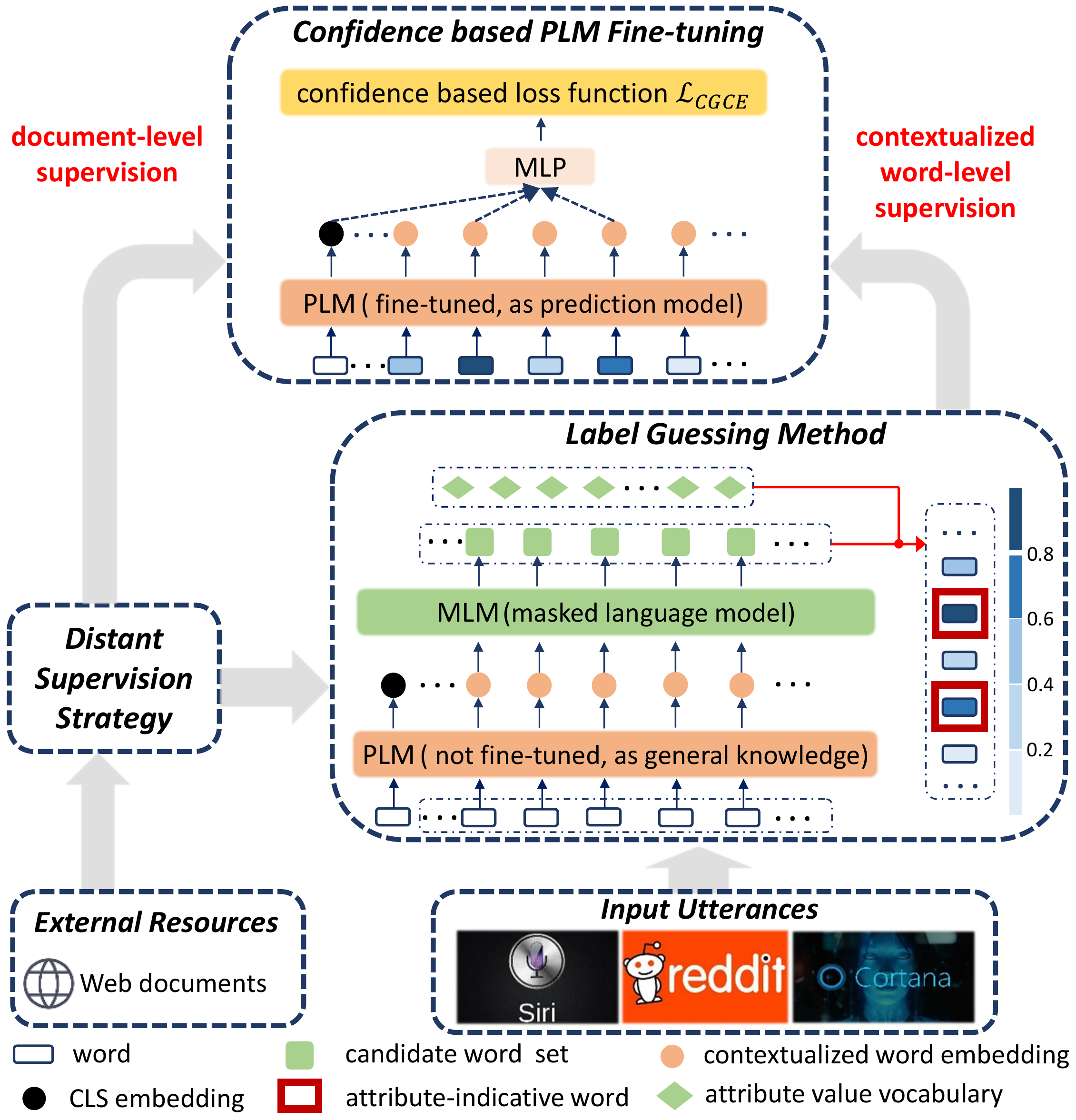}
	\caption{Overview of our framework.}
	\label{framework}
\end{figure}

\section{The framework: DSCGN}
The overview of our framework is shown in Figure
\ref{framework}.
We first introduce the distant supervision strategy and the label guessing method. Finally, we describe confidence based PLM fine-tuning.

\subsection{Distant Supervision Strategy}
\label{Distant Supervision}

To tackle the labeled utterances shortage issue, we leverage a distant supervision strategy to annotate external resources automatically to obtain document-level supervision. Intuitively, Wikipedia is beneficial for predicting personal attributes from conversations as a distant knowledge resource, since the semantics of personal attribute values (i.e., teacher, driver, and dancer) can be interpreted by their corresponding Wikipedia pages (Wiki-pages). Additionally, we found that some pages in Wikipedia about
concepts related to the attribute values (e.g., tools used for a profession and the profession’s
specializations) are also helpful for our task.
Accordingly, we could mine the attribute knowledge embedded in Wikipedia to help predict personal attributes from conversations. 
To derive high-quality labeled Wikipedia data, for each personal attribute value, we select its corresponding Wiki-page by string matching based on the surface forms of the attribute value (e.g., Dentist) and the Wiki-page title (e.g., Wiki: Dentist). 
To obtain more labeled Wikipedia data, we construct Wiki-category by identifying at least one relevant category for each
attribute value and including all leaf pages under
the category like \cite{tigunova2020charm}. In this way, the document-level supervision $S_{doc}=\{(d_i,A_{d_i})\}$ could be derived automatically, where $d_i$ denotes a Web document from external resources (e.g., the Wiki-page or a page in Wiki-category) and $A_{d_i}$ denotes the corresponding attribute value of $d_i$.


\subsection{Label Guessing Method}
\label{label guess}
When provided an attribute value, humans can understand its semantics based on
general knowledge by associating it with some attribute-indicative words. To create contextualized word-level attribute value supervision (contextualized word-level supervision) automatically, inspired by \cite{meng2020text}, we propose a label guessing method to find attribute-indicative words via the PLM pre-trained with the masked language model (MLM) task on large-scale text corpora. 

 First, for the $i$-th occurrence $w_i$ of the word $w$ in unlabeled utterances and external resources (e.g., Wiki-page and Wiki-category) provided by the distant supervision strategy, we feed its contextualized embedding $h_{{w}_{i}}$ produced by PLM encoder to the MLM head, which will output a probability distribution over
the entire vocabulary $V$, indicating the likelihood of each word $v\in V$ appearing at this position.	
\begin{equation*}
	p(v|h_{{w}_{i}})=Softmax(W_2\varphi\ (W_1h_{w_i}+b)),
\end{equation*}
 where $\varphi(\cdot)$ is the activation function; $W_1\in \mathbb{R}^{h\times h}, W_2\in \mathbb{R}^{|V|\times h}$, and $b\in \mathbb{R}^{h}$ are learnable parameters pre-trained with the MLM objective of PLM. Intuitively, words that are interchangeable most
 of the time are likely to have similar meanings. We retain top-$K$ (sorted by $p(v|h_{{w}_{i}})$) similar words as the candidate word set $C_{w_i}$ for the word occurrence $w_i$. Thus, we can generate $K$ valid replacement words for each occurrence of the attribute value. Subsequently, for an attribute value $a$, we form the attribute value vocabulary $V_{a}$ using the top-$M$ words ranked by frequency in all candidate word sets of its occurrences ($C_{a_1}$,$C_{a_2}$,$\dots$,$C_{a_n}$), discarding stopwords and words that appear in many attribute value vocabularies. Then, we define the semantic similarity $Sim_{w2a}(w_i,a)$ between word $w_i$ and attribute value $a$ as follows. 	
 \begin{equation*}
 Sim_{w2a}(w_i, a)=\frac{|C_{w_i}\cap V_a |}{|V_a|}
 \end{equation*}
 
 If this similarity is larger than a threshold $\beta$, $w_i$ is regarded as an attribute-indicative word with regard to the attribute value $a$.
 Thereby, from unlabeled utterances and external resouces, we obtain contextualized word-level supervision $S_{word}=\{(w_i, A_{w_i})\}$, where $w_i$ denotes an attribute-indicative word and $A_{w_i}$ denotes the corresponding attribute value of $w_i$.

\subsection{Confidence based PLM Fine-tuning}


A PLM can be fine-tuned by leveraging task-specific training data so that it can achieve a better result when performing that task. In this paper, to leverage the personal attribute knowledge embedded in both unlabeled utterances and external resources for predicting personal attributes from conversations, we try to fine-tune the PLM based on the document-level supervision $S_{doc}$ yielded by the distant supervision strategy (Section \ref{Distant Supervision}) and contextualized word-level supervision $S_{word}$ derived by the label guessing method (Section \ref{label guess}). 

Specifically, considering the balance between model convergence and noise-robustness, and in order to give more attention to the true attribute-indicative words and relieve the effect of the mislabeled words,
we propose a confidence based loss function $\mathcal{L}_{CGCE}$ based on the generalized cross entropy
(GCE) loss \cite{zhang2018generalized} by incorporating the confidence based weight function $\xi(\cdot)$ which is calculated by the semantic similarity $Sim_{w2a}(\cdot)$ as follows:


\begin{equation*}
	\mathcal{L}_{CGCE}=-\sum_{t\in {S_{doc}\cup S_{word}}}\xi{(t)}\frac{1-{Softmax(W_{A_t}h_t+b_{A_t})}^{q}}{q}
\end{equation*}

\begin{equation*}
	\xi{(t)}=\begin{cases}
		Sim_{w2a}(t, A_t) \ \ \ \text{if}\  t \ \text{refers to an attribute-indicative word} \\ \ \ \ \ \ \ \ \ \ \ \ \ \ \ \ \ \ \ \ \ \ \ \ \ \ \ \text{from} \ S_{word} \text{,}\\
		1 \text{\ \ \ \ \ \ \ \ \ \ \ \ \ \ \ \ \ \ \ \ \ \ \ \ if t refers to a Web document from}\  S_{doc} \text{.}
	\end{cases}
\end{equation*}
where $W_{A_t}$$\in$$\mathbb{R}^{K\times h}$ and $b_{A_t}$$\in$$\mathbb{R}^{K}$ are learnable parameters of the linear layer ($K$ is the number of attribute values); $0<q<1$ is a hyperparameter; If $t$ refers to an attribute-indicative word from $S_{word}$, $h_t$ denotes its contextualized embedding, otherwise if $t$ refers to a Web document from $S_{doc}$, $h_t$ denotes its CLS embedding.

After fine-tuning, given a user and a user attribute, we combine all utterances of a user as a sentence and give this sentence as the input of the fine-tuned PLM to obtain the probabilities of different attribute values of this attribute. 

\section{EXPERIMENTAL STUDY}

\subsection{Experimental Setting}
\label{setting}

\textbf{Data Sets.} Two real-world data sets (i.e., a profession data set and a hobby data set) are used in this experiment. They are annotated and  publicly provided by \cite{tigunova2020charm}, and extracted from publicly-available Reddit submissions and comments between $2006$ and $2018$. They both consist of about 6000 users, with a maximum of 500 and an average of 23 users per attribute value. The number of attribute values for profession (hobby) is $71$ ($149$) respectively.
Specially, for each Reddit user, all posts containing
explicit personal assertions used for labeling have been removed. 

\noindent \textbf{Evaluation Measures.} We adopt the same evaluation metrics MRR (Mean Reciprocal Rank) and nDCG (normalized Discounted Cumulative Gain) as previous works \cite{tigunova2020charm,tigunova2019listening}. 
The nDCG score is calculated by averaging all users. Considering that a user may have many attribute values, we follow the assumption that there is only one correct attribute value for each user. Thus, MRR can be calculated independently for each attribute value before averaging. 

\noindent \textbf{Baseline Models.} No-keyword + BM25 \cite{tigunova2020charm} utilizes a user’s full utterances and a Web document from $S_{doc}$ w.r.t. an attribute value as the input of BM25 \cite{robertson2009probabilistic}. BM25 is a strong unsupervised retrieval model to calculate the similarity between two documents.  
RAKE (TextRank) + BM25 (KNRM) \cite{tigunova2020charm} uses keywords extracted from a user’s full utterances by RAKE (TextRank) and a Web document from $S_{doc}$ w.r.t. an attribute value as the input of BM25 (KNRM). RAKE \cite{rose2010automatic} and TextRank \cite{mihalcea2004textrank} are two SOTA unsupervised keyword extraction methods. KNRM \cite{xiong2017end} is an efficient neural retrieval model, which considers semantic similarity via term embeddings.
BERT IR \cite{tigunova2020charm} splits both user utterances and a Web document from $S_{doc}$ into 256-token chunks respectively to fit the input size of BERT, and predicts
the relevance between them based on a binary cross-entropy loss. 
$\rm{HAM_{avg}}$ ($\rm{HAM_{CNN}}$, $\rm{HAM_{CNN-attn}}$, and $\rm{HAM_{2attn}}$) \cite{tigunova2019listening} can predict the score of different attribute values by training the neural network based on $S_{doc}$. $\rm{HAM_{CNN-attn}}$ and $\rm{HAM_{2attn}}$ both utilize attention mechanisms within and across documents from $S_{doc}$. $\rm{HAM_{avg}}$ and $\rm{HAM_{2attn}}$ both adopt two stacked fully connected
layers, while $\rm{HAM_{CNN}}$ and $\rm{HAM_{CNN-attn}}$ both adopt text classification CNN.
$\rm{CHARM_{BM25}}$ ($\rm{CHARM_{KNRM}}$) \cite{tigunova2020charm} uses keywords extracted from a user’s full utterances by BERT and a Web document from $S_{doc}$ as the input of BM25 (KNRM). 

\noindent \textbf{Setting Details.}
We utilize the BERT base-uncased model as the PLM. The training batch size and the learning rate are set to 4 and 1e-5 respectively. The parameters $K$, $M$, and $\beta$ of the label guessing method are set to $50$, $100$, and $0.4$ respectively. The hyperparameter $q$ is set to $0.4$. It is noted that the adopted lists of personal attribute values and the result for each baseline except $\rm{HAM_{avg}}$, $\rm{HAM_{2attn}}$, $\rm{HAM_{CNN}}$, and $\rm{HAM_{CNN-attn}}$,  are taken from \cite{tigunova2020charm} directly. To represent utterance embedding, the baselines \cite{tigunova2019listening} use word2vec embeddings pre-trained on the Google News corpus via MindSpore Framework\footnote{https://www.mindspore.cn/en}. 
Source code used in this paper is publicly available\footnote{https://github.com/CodingPerson/DSCGN}.

\begin{table}[!t]
	\centering
	\caption{Performance on predicting personal attributes from conversations using the external resource (Wiki-page or Wiki-category).}
	\scalebox{0.62}[0.67]{
		\begin{tabular}{|c|c|c|c|c|c|c|c|c|c|}
			\hline
			\multirow{3}{*}{\textbf{\shortstack{Labeled \\utterances}}} & \multirow{3}{*}{\textbf{Method}} & \multicolumn{4}{c|}{\textbf{Wiki-page}} & \multicolumn{4}{c|}{\textbf{Wiki-category}} \\
			\cline{3-10}          &       & \multicolumn{2}{c|}{\textit{\textbf{profession}}} & \multicolumn{2}{c|}{\textit{\textbf{hobby}}} & \multicolumn{2}{c|}{\textit{\textbf{profession}}} & \multicolumn{2}{c|}{\textit{\textbf{hobby}}} \\
			\cline{3-10}          &       & \multicolumn{1}{c}{MRR} & nDCG  & \multicolumn{1}{c}{MRR} & nDCG  & \multicolumn{1}{c}{MRR} & nDCG  & \multicolumn{1}{c}{MRR} & nDCG \\
			\hline
			\hline
			\multirow{10}[2]{*}{no} & No-keyword + BM25 & \multicolumn{1}{c}{.15} & .32  & \multicolumn{1}{c}{.16} & .42  & \multicolumn{1}{c}{.17} & .37  & \multicolumn{1}{c}{.13} & .35 \\
			& RAKE + BM25 & \multicolumn{1}{c}{.16} & .33  & \multicolumn{1}{c}{.17} & .42  & \multicolumn{1}{c}{.19} & .39  & \multicolumn{1}{c}{.14} & .37 \\
			& RAKE + KNRM & \multicolumn{1}{c}{.16} & .33  & \multicolumn{1}{c}{.12} & .32  & \multicolumn{1}{c}{.13} & .34  & \multicolumn{1}{c}{.12} & .31 \\
			& TextRank + BM25 & \multicolumn{1}{c}{.21} & .39  & \multicolumn{1}{c}{.21} & .46  & \multicolumn{1}{c}{.26} & .45  & \multicolumn{1}{c}{.20} & .42 \\
			& TextRank + KNRM & \multicolumn{1}{c}{.21} & .38  & \multicolumn{1}{c}{.15} & .36  & \multicolumn{1}{c}{.18} & .36  & \multicolumn{1}{c}{.16} & .36 \\
			& $\rm{HAM_{avg}}$ & \multicolumn{1}{c}{.06} & .07  & \multicolumn{1}{c}{.06} & .05  & \multicolumn{1}{c}{.06} & .07  & \multicolumn{1}{c}{.03} & .02 \\
			& $\rm{HAM_{2attn}}$ & \multicolumn{1}{c}{.06} & .07  & \multicolumn{1}{c}{.04} & .05  & \multicolumn{1}{c}{.06} & .07  & \multicolumn{1}{c}{.06} & .07 \\
			& $\rm{HAM_{CNN}}$ & \multicolumn{1}{c}{.20} & .18  & \multicolumn{1}{c}{.22} & .14  & \multicolumn{1}{c}{.27} & .34  & \multicolumn{1}{c}{.17} & .27 \\
			& $\rm{HAM_{CNN-attn}}$ & \multicolumn{1}{c}{.21} & .28  & \multicolumn{1}{c}{.13} & .10   & \multicolumn{1}{c}{.25} & .31  & \multicolumn{1}{c}{.16} & .25 \\
			& DSCGN & \multicolumn{1}{c}{\textbf{.43}} & \textbf{.57}  & \multicolumn{1}{c}{\textbf{.29}} & \textbf{.50}  & \multicolumn{1}{c}{\textbf{.44}} & \textbf{.60}   & \multicolumn{1}{c}{\textbf{.29}} & \textbf{.49} \\
			\hline
			\hline
			\multirow{3}[2]{*}{yes} & BERT IR & \multicolumn{1}{c}{.30} & .45  & \multicolumn{1}{c}{.22} & .43  & \multicolumn{1}{c}{.28} & .44  & \multicolumn{1}{c}{.18} & .42 \\
			& $\rm{CHARM_{BM25}}$ & \multicolumn{1}{c}{.29} & .46  & \multicolumn{1}{c}{.24} & .47  & \multicolumn{1}{c}{.28} & .47  & \multicolumn{1}{c}{.21} & .43 \\
			& $\rm{CHARM_{KNRM}}$ & \multicolumn{1}{c}{.27} & .44  & \multicolumn{1}{c}{.22} & .44  & \multicolumn{1}{c}{.35} & .55  & \multicolumn{1}{c}{.27} & \textbf{.49} \\
			\hline
	\end{tabular}}%
	\label{tab:addlabel}%
\end{table}%


\subsection{Experimental Results}
Besides external resources, the baselines BERT IR, $\rm{CHARM_{BM25}}$, and $\rm{CHARM_{KNRM}}$ require extra labeled utterances for model training, and perform ten fold cross-validation on the profession (hobby) data set under the zero-shot setting in which the attribute values in the training and test data are disjoint.
Our method DSCGN and other baselines 
do not require labeled utterances as input and perform on the profession (hobby) data set directly, which is a more difficult setting, since it pushes “zero-shot” to the extreme – no labeled utterances for any attribute values are given.

\noindent\textbf{Effectiveness Study.} From the results in Table \ref{tab:addlabel}, it can be seen that whichever external resource (i.e., Wiki-page or Wiki-category) is used, our proposed method DSCGN obtains the best performance compared with the twelve baselines on both data sets in terms of MRR and nDCG. To be specific, there are nine baselines that just leverage the external resources as distant supervision like us. Nevertheless, our method significantly outperforms them, which shows that our method can leverage the personal attribute knowledge embedded in external resources better. In addition to the external resources, the baselines $\rm{CHARM_{BM25}}$, $\rm{CHARM_{KNRM}}$, and BERT IR utilize extra labeled utterances for model training, which puts DSCGN at a disadvantage. However, in spite of this disadvantage, we found that our framework still promotes by at least about 9 (5) percentages compared with the best baseline in terms of MRR (nDCG) over the profession data set, and obtains the best performance in terms of MRR (nDCG) over the hobby data set, indicating the superiority of our proposed framework for the task of predicting personal attributes from conversations.


\noindent \textbf{Ablation Study.}
To show the different parts of our framework (i.e., confidence based weight function $\xi(\cdot)$, label guessing method, and distant supervision strategy) are crucial for our task, we conduct an ablation study by removing each part of the proposed framework DSCGN and testing the performance of the remaining parts over the profession (bobby) data set using Wiki-category (Wiki-page) as the external resource. From the experiment results shown in Table \ref{abla}, compared with these truncated versions of DSCGN, the full version of DSCGN obtains the best performance, which demonstrates each part of our framework is helpful for our task. To be specific, removing the label guessing method or the distant supervision strategy, the performance of DSCGN degrades a lot, indicating the personal attribute knowledge embedded in utterances and external resources are complementary.

\label{effectiveness}
\begin{table}[!t]
	\centering
	\caption{Ablation study.}
	\scalebox{0.7}{\begin{tabular}{|c|c|c|c|c|}
			\hline
			\multirow{3}{*}{\textit{\textbf{Ablations}}} & \multicolumn{2}{c|}{\textit{\textbf{profession}}} & \multicolumn{2}{c|}{\textit{\textbf{hobby}}} \\
			\cline{2-5}          & \multicolumn{2}{c|}{Wiki-category} & \multicolumn{2}{c|}{Wiki-page} \\
			\cline{2-5}          & MRR   & nDCG  & MRR   & nDCG \\
			\hline
			\makebox[0.1\textwidth][c]{DSCGN} & \makebox[0.06\textwidth][c]{\textbf{0.44}}  & \textbf{0.60}   &\makebox[0.06\textwidth][c]{\textbf{0.29}}  & \textbf{0.50} \\
			DSCGN w/o confidence based weight function $\xi(\cdot)$ & 0.42  & 0.58  & 0.27  & 0.48 \\
			DSCGN w/o label guessing method & 0.38  & 0.54  & 0.27  & 0.48 \\
			DSCGN w/o distant supervision strategy & 0.06  & 0.23  & 0.04  & 0.20 \\
			\hline
	\end{tabular}}%
	\label{abla}%
\end{table}%

\section{related work}
\textbf{Personal Attribute Prediction from Conversations.}
Previous works \cite{jing2007extracting,li2014personal} relied on the condition that user attributes are explicitly mentioned in the user utterances, so these methods are not suitable for the profession (hobby) data set without personal assertions used in this paper.
Recently, HAM \cite{tigunova2019listening} proposed some models based on neural network architectures (e.g., stacked fully connected layers and CNNs), to predict a score of different attribute values (e.g., different professions) for a given subject-predicate pair by using average methods or attention mechanisms within and across utterances. CHARM \cite{tigunova2020charm} first leveraged the pre-trained LM BERT to extract keywords related to the attribute from user utterances, and then adopted some SOTA ranking models based on information retrieval techniques (i.e., RAKE and KNRM) to match these keywords against documents that indicate possible values of the attribute. Specially, they proposed a reinforce policy gradient method to train the process of keywords extraction.

\noindent\textbf{Personal Attribute Prediction from Social Media.}
In recent years, numerous methods
have been proposed to predict attributes (e.g., age \cite{bayot2017age,liu2021age}, gender \cite{bayot2017age,mac2017demographic}, and occupational class \cite{preoctiuc2015analysis}) of users from social media (e.g., Twitter and Facebook). The vast majority of these works rely on rich meta-data of social media such as hashtag, user profile description, and social
network structure. This makes them unsuitable for our task, as conversations do not have these rich meta-data. In addition, there are only three works \cite{basile2017n,bayot2017age,preoctiuc2015analysis} that predict user latent attributes based solely on user-generated text. However, from the results provided by \cite{tigunova2019listening}, we found that these three methods perform not well on the task of predicting personal attributes from conversations, so we did not select them as our baselines in this paper. Additionally, some recent works explored the task of named entity location prediction from Twitter \cite{shen2018predicting,liu2020named}, but their methods are not applicable for personal attributes.


\section{Conclusion}
To tackle three issues existing in previous studies (i.e., labeled utterances shortage, personal attribute knowledge underutilization, and unsatisfactory performance),
we propose a flexible PLM based framework DSCGN without requiring any labeled utterances. Two categories of supervision (i.e., document-level supervision and contextualized word-level supervision) are utilized by DSCGN to fine-tune the language model via a noise-robust loss function. To demonstrate the
effectiveness of DSCGN, we conduct extensive experiments over two real-world data sets, which show whichever external resource (i.e., Wiki-page or Wiki-category) is used, our framework obtains the best performance compared with all the twelve baselines in terms of nDCG and MRR.

\begin{acks}
This work was supported in part by National Natural Science Foundation of China (No. U1936206), Natural Science Foundation of Tianjin (No. 19JCQNJC00100), YESS by CAST (No. 2019QNRC001), and CAAI-Huawei MindSpore Open Fund. 
\end{acks}

\bibliographystyle{ACM-Reference-Format}
\bibliography{sample-base}
\end{document}